 \documentclass[final,5p,times,twocolumn]{elsarticle}
\usepackage{amssymb}
\usepackage{graphics} 
\usepackage{epsfig} 
\usepackage{color}
\usepackage{makecell}
\usepackage{multirow}
\usepackage{textcomp}
\usepackage{color}
\usepackage{diagbox}
\usepackage{upgreek}
\usepackage{textcomp}
\usepackage{subfigure}
\usepackage{algorithm}  
\usepackage{algorithmic}  
\usepackage{booktabs}
\usepackage{amsfonts}
\usepackage{lineno,hyperref}
\usepackage{amsmath}

\journal{Neurocomputing}

\biboptions{sort&compress}

\begin{document}

\begin{frontmatter}



\title{A novel channel pruning method for deep neural network compression}


\author[]{Yiming Hu}
\ead{huyiming2016@ia.ac.cn}

\author[]{Siyang Sun}
\ead{sunsiyang2015@ia.ac.cn}

\author[]{Jianquan Li}
\ead{lijianquan2015@ia.ac.cn}

\author[]{Xingang Wang}
\ead{xingang.wang@ia.ac.cn}

\author[]{Qingyi Gu\corref{cor1}}
\ead{qingyi.gu@ia.ac.cn}
\cortext[cor1]{Corresponding author}

\address{Research Center of Precision Sensing and Control, Institute of Automation, Chinese Academy of Sciences, Beijing 100190, China}
\address{School of Computer and Control Engineering, University of Chinese Academy of Sciences, Beijing 101408, China}

\begin{abstract}
In recent years, deep neural networks have achieved great success in the field of computer vision. However, it is still a big challenge to deploy these deep models on resource-constrained embedded devices such as mobile robots, smart phones and so on. Therefore, network compression for such platforms is a reasonable solution to reduce memory consumption and computation complexity. In this paper, a novel channel pruning method based on genetic algorithm is proposed to compress very deep Convolution Neural Networks (CNNs). Firstly, a pre-trained CNN model is pruned layer by layer according to the sensitivity of each layer. After that, the pruned model is fine-tuned based on knowledge distillation framework. These two improvements significantly decrease the model redundancy with less accuracy drop. Channel selection is a combinatorial optimization problem that has exponential solution space. In order to accelerate the selection process, the proposed method formulates it as a search problem, which can be solved efficiently by genetic algorithm. Meanwhile, a two-step approximation fitness function is designed to further improve the efficiency of genetic process. The proposed method has been verified on three benchmark datasets with two popular CNN models: VGGNet and ResNet. On the CIFAR-100 and ImageNet datasets, our approach outperforms several state-of-the-art methods. On the CIFAR-10 and SVHN datasets, the pruned VGGNet achieves better performance than the original model with 8$\times$ parameters compression and 3$\times$ FLOPs reduction.

\end{abstract}


\end{frontmatter}


\section{Introduction}
In the past few years, CNN models have been widely applied to various computer vision tasks, e.g., image classification, object detection, action recognition, since AlexNet \cite{Krizhevsky2012} won the ImageNet Challenge: ILSVRC 2012~{\cite{Russakovsky2015}}. Recently, for pursuing better accuracy, to design deeper and wider CNN models has become a general trend, such as VGGNet \cite{Simonyan2014}, ResNet~{\cite{He2015}} and Xception~{\cite{Chollet2016}}.

However, it is difficult to deploy these deep models on resource-constrained devices including mobile robots, unmanned aerial vehicles, smart phones, etc. On the one hand, convolution operations exhaust huge computation resources and require adequate power, which are scarce on mobile devices. On the other hand, billions of network parameters are also high storage overhead for embedded devices. Take the VGG-16 model as an example, it has over 138 million parameters and occupies more than 500MB memory space. Meanwhile, 30 billion float-point-operations (FLOPs) are needed to classify a $224\times224$ image. Obviously, it is impractical to deploy such large model into embedded devices directly.

Therefore, it is a critical problem to compress deep models without significant accuracy drop. In view of this, many model compression and acceleration methods have been proposed recently including weight pruning \cite{Han2015, Guo2016, Hu2016, Luo2017, He2017}, network quantization \cite{Rastegari2016, Cour2016, LiZ2017}, low-rank approximation \cite{Denton2014, Wang2016}, and efficient model designs \cite{Zhang2017, Howard2017, Xie2017}. Neuron pruning methods \cite{Han2015, Guo2016} make weights sparse by removing some less important connections. Network quantization \cite{Cour2016} is proposed for storage space compression by decreasing the presentation precision of parameters. However, these two methods require special software or hardware implementations for acceleration. Low-rank approximation \cite{Denton2014} decomposes weight matrices into several small ones with less storage by means of low-rank matrix techniques, which is not efficient for those $1\times1$ convolutions. While efficient model designs \cite{Zhang2017, Howard2017} focus more on acceleration instead of compression by optimizing convolution operations or network architectures.

Channel pruning \cite{Hu2016, Luo2017, He2017} is another type of weight pruning method, which is different from neuron pruning. Compared with removing single neuron connection, pruning the whole channel has two advantages. Firstly, it does not introduce sparsity to the original network structure, thus requires no special software or hardware implementations for the resulting models. Secondly, it does not require huge disk storage and run-time memory in inference stage. Recently, some training-based channel pruning methods \cite{Liu2017, Wen2016} by adding regularization terms to weights in training stage have been proposed. Nevertheless, training from scratch is very time-consuming especially for some large datasets such as ImageNet~{\cite{Deng2009}}. To ignore the pre-training process, many works \cite{Luo2017, He2017} are presented to perform channel pruning on pre-trained models with different pruning criteria. However, these existing methods still have much space for improvement in decreasing model redundancy. Furthermore, most works \cite{Liu2017, Wen2016, Luo2017, He2017} only accelerate networks in inference stage and few of them pay attention to the off-line pruning efficiency.

In this paper, a new channel pruning method based on genetic algorithm is proposed to compress and accelerate deep CNN models. Three main contributions of this work are as follows:  
\begin{enumerate}
	\item A channel selection strategy based on genetic algorithm is presented for exploring a compact network architecture. Channel selection is considered as a combinatorial optimization problem which has exponential solution space, and genetic algorithm is proved to be an effective measure to deal with this issue. 
	\item A two-step approximation fitness function is designed to ensure the searching efficiency of genetic algorithm through specific genetic operations.
	\item A common way of network fine-tuning based on knowledge distillation framework is proposed, which is experimentally demonstrated effective.
\end{enumerate}

Experiments are conducted on three benchmark datasets with two popular CNN models: VGGNet and ResNet. On the CIFAR-100~{\cite{Krizhevsky2009}} and ImageNet datasets, our method outperforms several state-of-the-art channel selection strategies \cite{Luo2017, Li2016, Molchanov2016}. On the CIFAR-10~{\cite{Krizhevsky2009}} and SVHN~{\cite{Netzer2011}} datasets, the pruned VGGNet and ResNet achieve higher accuracy than the original model with 8$\times$ parameters compression and 3$\times$ FLOPs reduction.

This paper is organized as follows: section II introduces related works in model compression and acceleration. Section III describes the framework and details of our approach. The experimental results are illustrated and discussed in section V. Section VII gives conclusion and future works.

\begin{figure*}[t]
	\centering
	\includegraphics[width=1.5\columnwidth]{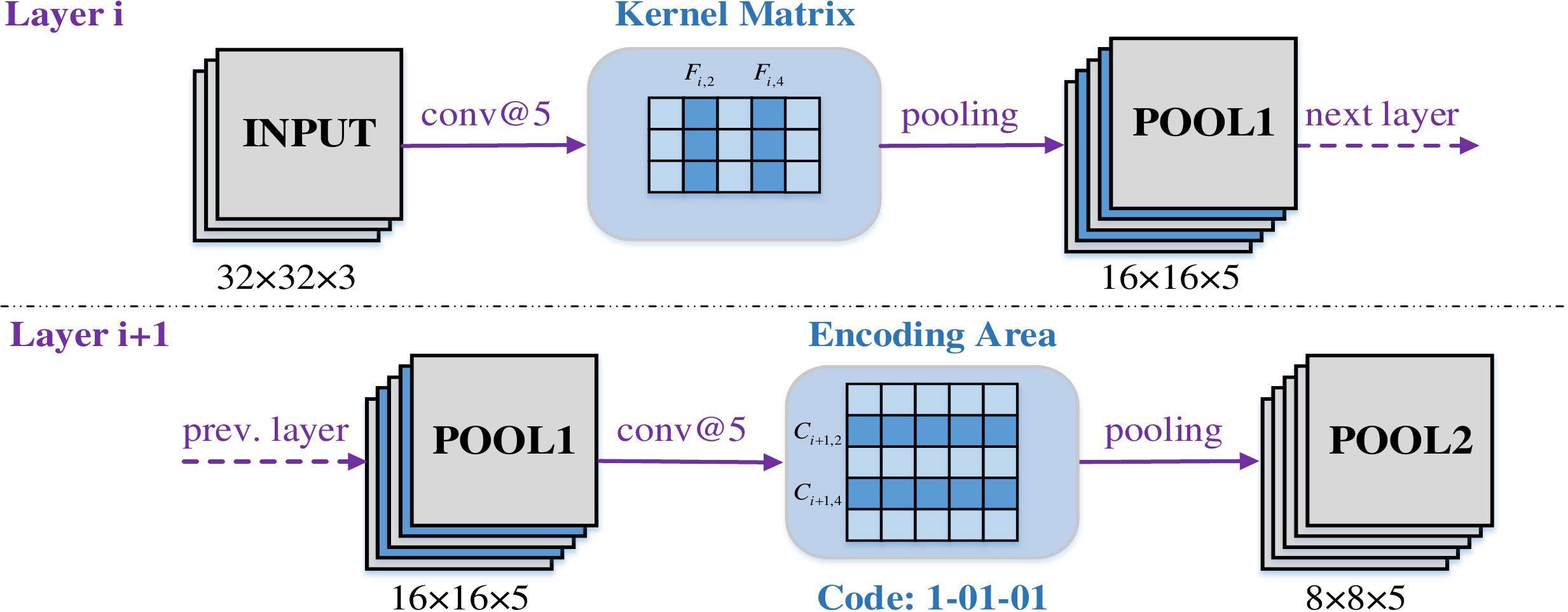}
	\vskip -2mm
	\caption{{the binary encoding method for our channel pruning strategy. This figure shows that two channels are pruned for convolution kernels in layer $i+1$. Thus relevant channels of feature map and filters in layer $i$ are also removed simultaneously, which does not introduce sparsity to the original network. Our optimization algorithm works in the encoding area. $F_{i,2}, C_{i+1,4}$ : second filter in $i$-th layer and fourth channel in layer $i+1$.}}
	\label{fig:EncodingMethod}
\end{figure*}

\section{Related Works}

\subsection{Model compression and acceleration}

In recent years, many researchers have made effort in the field of network compression and acceleration, and plenty of algorithms have been presented, that can be divided into four types.

\subsubsection{Weight pruning}
Han et al.~{\cite{Han2015}} introduced a simple pruning method: weights lower than a certain threshold are considered as low contribution ones, that can be pruned, and then fine-tuning is took for restoring the network accuracy. This process is executed iteratively, until a sparse model is generated. However, runtime memory and inference time of the pruned model with sparse weights will not reduce in contrast with the baseline model, unless special software or hardware implementations are adopted. In view of this issue, Structured Sparsity Learning (SSL)~{\cite{Wen2016}} added group sparsity regularizations to different level of structures such as filters, channels or layers, then some weights belonging to the same level of structure would go to zero simultaneously, that could be removed. This method does not introduce sparsity to the original model but bring extra overhead to the training process. To avoid this problem, Liu et al.~{\cite{Liu2017}} imposed L1 regularization on the batch normalization (BN) scaling factors directly. However, training from scratch is still time-consuming. Afterwards, some inference-based channel pruning methods \cite{Hu2016, Luo2017, He2017} were proposed, and the core of these methods is how to define selection criteria. Li et al.~{\cite{Li2016}} believed those filters with smaller weights always produce weaker activations, thus they could be removed. But the criterion may remove some important filters especially at shallow layers. Hu et al.~{\cite{Hu2016}} evaluated the importance of each channel based on its sparsity and removed those channels whose output activations contain more zero values, which shows poor performance at convolution layers. Molchanov et al.~{\cite{Molchanov2016}} used the first-order Taylor expansion to approximate the loss change of object function, then removed those channels which have less influence on the loss function. This method pays more attention to model acceleration, thus its performance is limited. Thinet~{\cite{Luo2017}} was proposed to prune filters based on statistics information calculated from its next layer via greedy algorithm. But greedy algorithm would not be the best way of solving the combinatorial optimization problem especially for relatively large solution space. Furthermore, its off-line pruning process is very time-consuming, since it need to traverse the entire training sets at each iteration step.

\subsubsection{Network quantization} 
In order to reduce the model size, HashNet~{\cite{Chen2015}} mapped the network parameters into different groups and each group shared the same values. Deep compression~{\cite{Hansong2015}} compressed the VGG-16 model from 552MB to 11.3MB without accuracy drop. However, these two methods only focus on practical storage size instead of runtime memory. Courbariaux et al. proposed BinaryNet \cite{Cour2016} whose weights were quantized to be $+1$ or $-1$, which achieved a state-of-the-art result on the MNIST and CIFAR-10 datasets. To further reduce the computation cost, XNOR-Net~{\cite{Rastegari2016}} was proposed to use both binary weights and inputs, yet achieved a poor performance on the ImageNet dataset. To improve accuracy, High-Order Residual Quantization (HORQ)~{\cite{LiZ2017}} performed convolution operations on inputs in different scales and then combined the results. The accuracy gap shrinks a lot compared with XNOR-Net, which is still unacceptable to practical applications.

\subsubsection{Low-rank approximation}
Singular Value Decomposition (SVD) is a popular low-rank matrix decomposition measure. Denton et al.~{\cite{Denton2014}} performed elementary tensor decomposition based on SVD to approximate weight matrix, to exploit the model redundancy and reduce computation overhead. To further reduce the model redundancy, Wang and Cheng~{\cite{Wang2016}} proposed a Block-Term Decomposition (BTD) method based on low-rank and group sparse decomposition. These two methods have quite good performance mainly on FC (fully-connected) layers, achieving 3$\times$ compression without significant FLOPs reduction. However, most of computation is distributed at convolution layers.

\subsubsection{Efficient model designs}
High demand on embedded devices stimulates efficient network structure designs. For instance, GoogleNet~{\cite{Szegedy2015}} was proposed to use inception module to reduce computation cost instead of simply stacking many layers. ResNet achieves quite good performance by introducing efficient bottleneck structure. MobileNet~{\cite{Howard2017}} adopted depth-wise convolution and obtains state-of-the-art results. ShuffleNet~{\cite{Zhang2017}} generated the idea of group and depth-wise convolution, which achieved obvious acceleration on ResNet. ResNeXt~{\cite{Xie2017}} explored the split-transform-merge strategy on ResNet and obtained better results than the baseline. However, these methods pay more attention to acceleration instead of compression, thus the model size is not reduced significantly.

\subsection{Genetic algorithm}

Genetic algorithm simulates the artificial population evolution process to solve optimization problems \cite{Beasley1996, Deb2002}. The core idea of genetic algorithm is genetic operators including crossover and mutation, which allows us to explore larger solution space. The mutation operator is used to maintain genetic diversity and avoid local minima when producing the next generation. The crossover operator is beneficial to exchange genes among different individuals. Genetic algorithm has been widely applied in various fields such as function optimization \cite{Pravesjit2017}, self-adaptation control \cite{Patrascu2017}, pattern recognition \cite{Abed2010}, etc. Genetic CNN \cite{XieL2017} used genetic algorithm to find efficient deep network structure automatically. In this situation, each network structure is encoded in a fixed-length binary series, and the fitness function is defined by final classification accuracy. There are lots of works to study how to improve the performance of genetic algorithm such as performing local search \cite{Ulder1990} and generating random keys \cite{Snyder2006}. Different from previous works \cite{Bayer2009, Ding2013} which used genetic algorithm to explore deep neural network architectures, our method aims to get a compact network by performing pruning on a pre-trained model.

\begin{algorithm*}[h]
	\caption{Genetic algorithm for single layer pruning}
	\begin{algorithmic}[1]
		
		\label{alg:genetic_algorithm}
		\STATE \textbf{Initialization}: let $t=0$; generate the initial population $\{\mathbb{I}_{m,n}\}_{n=1}^N$; initialize the crossover probability $p_C$, the mutation probability $p_M$, the number of individuals of population, and the maximum iteration number $T$. 
		\STATE \textbf{Input:} the Hessian matrix and parameters of the layer to be pruned, the compression rate $r$ and the train sets $\{x_i, y_i\}$.
		\STATE \textbf{Output:} a set of binary series encoding the desired convolution kernel structures with highest fitness.
		\WHILE {$t < T$}
		\STATE \quad\textbf{Evaluation}: calculate the fitness of each individual in $t$-th generation $\{\mathbb{I}_{m,n}\}_{n=1}^N$ with Eq.~({\ref{eq:second_order_term}});
		\STATE \quad\textbf{Selection}: select individuals with high fitness from $\{\mathbb{I}_{m,n}\}_{n=1}^N$ to $\{\mathbb{I}_{m+1,n}\}_{n=1}^N$ by the Roulette Wheel Algorithm;
		\STATE \quad\textbf{Crossover}: select two individuals \{$\mathbb{I}_{m+1,p}$, $\mathbb{I}_{m+1,q}$\} randomly, then take on crossover operation with probability $p_C$;
		\STATE \quad\textbf{Mutation}: select one non-crossover individual \{$\mathbb{I}_{m+1,r}$\} randomly, then do mutation operation with probability $p_M$;
		\STATE \quad Update $t = t + 1$;
		\ENDWHILE

	\end{algorithmic}
\end{algorithm*}

\section{Our Approach}

\subsection{Encoding method}
In order to encode a convolution kernel into a chromosome, it is represented with a binary series whose length is equal to the number of channels in the convolution kernel. Each channel is encoded with 0 or 1, and then the channels with code 0 are simply pruned. 

The initial population is a set of simplest possible convolution kernels $\{\mathbb{K}_n\}_{n=1}^N$. $\mathbb{K}_n$ is a binary series, i.e., $\mathbb{K}_n \in \{0,1\}^C$,  and each bit in $\mathbb{K}_n$ obeys the Bernoulli distribution: $\mathbb{K}_n^c \sim \mathcal{B}(p) $, $c = 1,2,3, ..., C$, where, p is the expected value of Bernoulli random variable and C is the number of channels. In our experiments, p is simply set to the proportion of the channels that need to be preserved, which is beneficial to the fast convergence of genetic algorithm. After that, we evaluate each individual in the initial population with the fitness function (described in Section \ref{fitness_function}) and then choose fitter ones according to their fitness to do genetic operations, so as to find competitive structures in the whole solution space.

Fig.~{\ref{fig:EncodingMethod}} shows an example of the convolution kernel encoding. However, there are $2^n$ possible structures for a convolution kernel with n channels after channel pruning. It's NP hard to find the optimal one from the entire solution space, thus genetic algorithm is adopted to solve the problem. 

\subsection{Genetic operators}

We first evaluate each individual of $n$-th generation and then choose fitter ones according to their fitness, so that the offspring produced in the next generation have higher fitness. Roulette Wheel algorithm is adopted as our selection strategy. The algorithm flow is as follows: first, $n$-th individual of $m$-th generation is defined as $\mathbb{I}_{m,n}$ and then is assigned a fitness $f_n$. The probability of being selected for $\mathbb{I}_{m,n}$ is the proportion of $f_n$ in the fitness sum of all individuals from $m$-th generation, which means that the larger the fitness of individual $\mathbb{I}_{m,n}$ is, the higher probability of being selected it has. Then, each individual of the next generation is produced from $\{\mathbb{I}_{m,n}\}_{n=1}^N$. In the process of evolution, the best found individual is expected to be preserved. In order to prevent the individual with largest fitness from being lost when producing the new generation via crossover and mutation, elitism which first copies the best individual to the next generation without reproduction is adopted in the selection operator.

The mutation operator is used to maintain genetic diversity and avoid local minima when producing the next generation. A random bit is first chosen from individual $\mathbb{I}_{m,n}$ and the probability of being selected for each bit is ${1}/{l}$, where $l$ is the length of binary coding series. Then the chosen bit is inverted according to a small probability $p_M$, i.e., if the chosen bit is 0, it is changed to 1 with the probability of $p_M$ and vice versa. The crossover operator produces new offspring from two selected parents $\mathbb{I}_{m,p}$ and $\mathbb{I}_{m,q}$, which contributes to gene flow in the entire population. Single point crossover is used in crossover operator. Just as mutation operator, one crossover bit is first selected with the same probability as mutation, then the binary series from beginning to the crossover point is copied from $\mathbb{I}_{m,p}$ and the rest is copied from $\mathbb{I}_{m,q}$ with a small probability $p_C$. In our experiment, $p_M$ and $p_C$ are both set to 0.1. If they are set too large, genetic process will become random search. After crossover and mutation, a new individual is generated. Each individual of the next generation is produced in the same way. 

\subsection{Improved fitness function}
\label{fitness_function}
Before selection operation, each individual of $n$-th generation is assigned a fitness function to determine whether it will survive. The selection of fitness function correlates quite highly with the quality of final solution. Our ultimate goal is to get a compact network with good classification performance, so an ideal fitness function is the classification accuracy of the network. However, the evaluation process is performed repeatedly in genetic algorithm, thus need to be sufficiently fast. For this reason, an improved fitness function is designed from three aspects to reduce computation complexity .

\subsubsection{Layer-wise error function} 
A convolution filter is denoted as $W\in \mathbb{R}^{C\times K \times K}$. $X\in \mathbb{R}^{N\times C\times K \times K}$ represents the volumes sampled from input feature map. $X$ and $W$ produce an N-dimensional output vector Y, where $N$ is the number of samples. Then a layer-wise error function is designed as:
\begin{gather}
E(\widetilde{Y}) = \frac{1}{N}\sum_{i=1}^N||\widetilde{Y}_i-Y_i||^2
\label{eq:layer_wise_error_function}
\end{gather}
where $\widetilde{Y}$ is the output before performing activation function after channel pruning. For each individual $\mathbb{I}_{m,n}$, its error is calculated with Eq.~({\ref{eq:layer_wise_error_function}}) and treat the error as fitness. Therefore, the target of genetic algorithm is to find the channels whose deletion can minimize the error function. Layer-wise error function only involves single layer forward computation instead of entire network inference, which is relatively efficient in contrast with computing the classification accuracy. However, it is still computationally expensive to traverse all training samples for each individual when the training set is very large. A feasible scheme is to approximate Eq.~({\ref{eq:layer_wise_error_function}}) via second-order Taylor expansion.

\begin{table*}[t]
	\small
	\renewcommand\arraystretch{1.2}
	\caption{performance of several layers with different pruned rate. First 12 layers of VGG-16 model are pruned respectively on the ImageNet dataset with different compression rate without fine-tuning. The data unit in the table is percentages. }
	\label{table:sensitivety}
	\begin{center}
		\begin{tabular}{ccccccccccccc}
			\toprule
			\makecell[c]{Compression rate} & \makecell[c]{conv1} & \makecell[c]{conv2} & \makecell[c]{conv3} & \makecell[c]{conv4} & \makecell[c]{conv5}& \makecell[c]{conv6}& \makecell[c]{conv7}& \makecell[c]{conv8}& \makecell[c]{conv9}& \makecell[c]{conv10}& \makecell[c]{conv11}& \makecell[c]{conv12}\\ 
			\midrule
			\makecell[c]{20\%}& 69.02 & 67.66 & 69.01 & 65.31 & 66.57 & 66.72 & 64.85 & 66.07 & 66.19 & 65.33 & 64.26 & 64.67 \\
			\makecell[c]{40\%}& 69.01 & 63.48 & 65.08 & 58.98 & 58.76 & 57.49 & 51.69 & 55.36 & 54.88 & 55.56 & 50.14 & 51.44 \\
			\makecell[c]{60\%}& 68.07 & 50.24 & 50.86 & 34.33 & 31.49 & 32.01 & 22.98 & 20.67 & 21.24 & 20.74 & 21.08 & 29.59 \\
			\makecell[c]{80\%}& 50.72 & 21.52 & 10.05 & 9.22 & 7.72 & 5.36 & 6.14 & 6.76 & 9.35 & 7.71 & 4.11 & 3.32 \\
			\bottomrule	
		\end{tabular}
	\end{center}
\end{table*}

\subsubsection{Second-order Taylor Approximation} 
As shown in~{\cite{Lecun1989}}, Eq.~({\ref{eq:layer_wise_error_function}}) can be approximated as follows:
\begin{gather}
\delta E = (\frac{\partial E}{\partial W})^T \delta W + \frac{1}{2}\delta W^TH\delta W + O(||\delta W||^3)
\label{eq:second_order_expansion}
\end{gather}

where $\delta E$, $\delta W$ represent the perturbation of objective function and parameters,  $H$ is the Hessian matrix w.r.t $W$, and $O(||\delta W||^3)$ denotes high-order terms. In Optimal Brain Damage(OBD)~{\cite{Lecun1989}}, a pre-trained network is thought to be at a local minimum, thus the first order term can be ignored. For convenience, the high-order terms are also neglected. Now, Eq.~({\ref{eq:layer_wise_error_function}}) can be further simplified as:
\begin{gather}
\delta E = \frac{1}{2}\delta W^TH\delta W
\label{eq:second_order_term}
\end{gather}

Eq.~({\ref{eq:second_order_term}}) can be also regarded as the sensitivity of parameters $W$. Channel pruning based on genetic algorithm is translated to find a set of low-sensitivity channels and remove them.

\subsubsection{Approximation of the Hessian Matrix}
Eq.~({\ref{eq:second_order_term}}) involves calculating the Hessian matrix, which is computationally intractable. Existing method in~{\cite{Dong2017}} is adopted to get an approximate Hessian matrix. According to the chain rule, the first derivative of the error function w.r.t $W$ is:
\begin{gather}
\frac{\partial E}{\partial W} = -\frac{1}{N}\sum_{i=1}^{N}\frac{\partial Y_i}{\partial W}(\widetilde{Y}_i-Y_i)
\label{eq:first_order_grad}
\end{gather}
where $\widetilde{Y}$ is considered as a constant independent of weight $W$, since it is the output after channel pruning. Note that the constant coefficient in Eq.~({\ref{eq:first_order_grad}}) has been simply ignored. According to the definition of the Hessian matrix, $H$ can be denoted as:
\begin{gather}
\frac{\partial^2 E}{\partial (W)^2} = \frac{1}{N}\sum_{i=1}^{N}(\frac{\partial Y_i}{\partial W}(\frac{\partial Y_i}{\partial W})^T-\frac{\partial^2 Y_i}{\partial (W)^2}(\widetilde{Y}_i-Y_i)^T)
\label{eq:second_order_grad}
\end{gather}
In practice, $y$ is close to $\widetilde{y}$ in most cases and computing second-order derivative is time-consuming, thus the terms including $\widetilde{Y}_i-Y_i$ is simply neglected. Meanwhile, the derivative of $Y_i$ with respect to $W$ is exactly equal to the input $X_i$. Therefore, the Hessian matrix can be acquired via Eq.~({\ref{eq:hessian_matrix}}). More importantly, the Hessian matrix of each filter at the same layer is identical,  which significantly reduces the computation cost. 
\begin{gather}
H = \frac{1}{N}\sum_{i=1}^{N}X_iX_i^T
\label{eq:hessian_matrix}
\end{gather}

To conclude, the prerequisite of computing the Hessian matrix is to get the input volumes $X$ from the input feature map of current layer by the pre-trained network inference. Then $H$ can be obtained via Eq.~({\ref{eq:hessian_matrix}}). Here, $X_i$ is sampled from different images and spatial locations.

\begin{figure}[t]
	\centering
	\includegraphics[width=1.0\columnwidth]{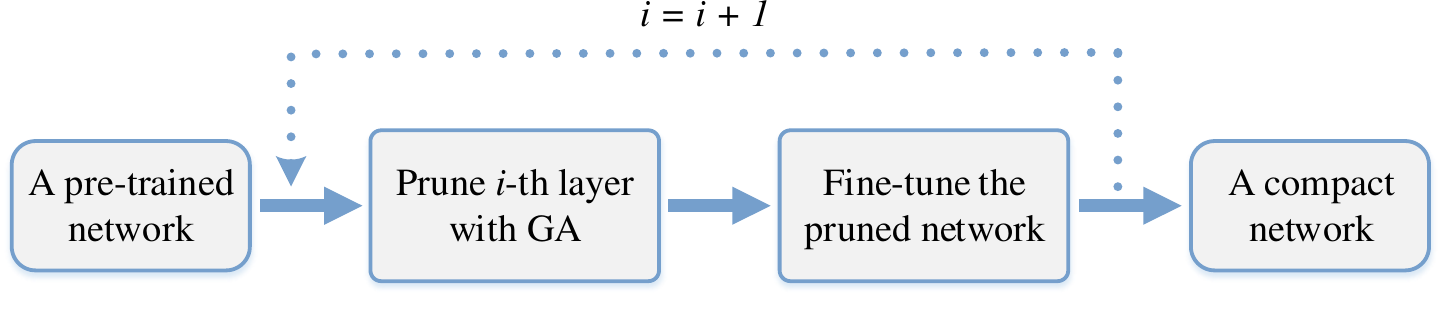}
	\vskip -2mm
	\caption{{flow chart of whole model pruning. The model is pruned layer by layer sequentially.}}
	\label{fig:PruningFlowchart}
\end{figure}

\subsection{Pruning whole model}
\label{PruningWholeModel}

The above genetic process is used to prune single layer, which is illustrated in Algorithm~{\ref{alg:genetic_algorithm}} . After pruning each layer, the pruned model is fine-tuned for several epochs to further restore the model accuracy. One or two epochs are used for the previous layers, while more epochs are adopted for the last layer. Next, Algorithm~{\ref{alg:genetic_algorithm}} is extended to the whole model pruning. Following~{\cite{Zhang2016}}, the whole model is pruned layer by layer sequentially. The process is described in Fig.~{\ref{fig:PruningFlowchart}}. However, we found that the redundancy of each layer is different, so the compression rate for each layer should not be identical. The sensitivity of each layer is defined and it depends on the accuracy change when pruning this layer with a certain compression rate. The layers with smaller accuracy drop have lower sensitivity, thus are pruned more aggressively. 

Take VGG-16 model on the ImageNet dataset as an example. As shown in Table \ref{table:sensitivety} , pruning at some sensitive layers (e.g., conv7, conv9) leads to a sharp drop in accuracy, thus each layer is assigned a compression rate according to its sensitivity. The deeper layers show higher sensitivity, thus can be pruned more aggressively. To make this idea feasible, all layers of VGG-16 model are coarsely divided into four groups according to their sensitivity. The first and second group are comprised of conv1 to conv3 and conv4 to conv6 respectively. The third group is composed of conv7 to conv10, and conv11 to conv12 fall into the last group. Then each group shares the same compression rate.

\subsection{Fine-tuning based on knowledge distillation}
\label{KnowledgeDistill}
In order to improve the fine-tuning process, knowledge distillation framework is introduced. We regard the pruned model as student model denoted as $S$, and treat original model as teacher model denoted as $T$. The teacher model is used to help restore the accuracy of student model by transferring knowledge from $T$ to $S$. Now, the goal of fine-tuning for student model is not only to make correct predictions but also to have the feature maps that are similar to the teacher's. However, the number of channels of student model is not the same as the teacher's after pruning. So attention map~{\cite{Zagoru2016}} is adopted to deal with the issue. Attention map is actually the mean of original feature map at channel dimension. Let $L(W,x)$ denote a standard cross entropy loss. Let $I$ denote the indices of teacher-student attention map pairs that need to be transfered. Then the following loss for student model can be defined as: 
\begin{gather}
L_{AT} = L(W_S,x) + \beta\sum_{j\in I}||F_T^j-F_S^j||
\end{gather}
where $\beta$ controls the ratio of two losses. $F_T$ and $F_S$ represent the attention maps from teacher and student respectively.

\section{Experiments}
In this section, the performance of our approach is empirically evaluated with the popular VGG-16 and ResNet models on the CIFAR, SVNH, and ImageNet datasets.

\subsection{Datasets and models}

\subsubsection{CIFAR}
CIFAR-10 contains 50,000 training images and 10,000 test images spanning 10 categories of objects, while CIFAR-100 have identical number of images spanning 100 categories of objects. The resolution of each image is 32$\times$32. The same data augmentation operations as~{\cite{He2016}} is used in training and fine-tuning stage. To verify the effectiveness of our approach, it is compared with different methods \cite{Luo2017, Li2016, Molchanov2016} on the CIFAR dataset.

\subsubsection{SVHN}
SVHN incorporates over 600,000 digital images which are obtained from house numbers in Google Street View images, and the size of these images are all 32$\times$32. For data augmentation, the input images are first normalized, then random horizontal flip is adopted. We prune VGGNet and ResNet with different compression rate on the SVHN dataset.

\subsubsection{ImageNet} ImageNet consists of 1.26 million training images and 50,000 test images spanning 1,000 categories of objects. In training and fine-tuning stages, all images are first resized to 256$\times$256, then randomly cropped into 224$\times$224, finally random horizontal flip is adopted. In test stage, all models are evaluated on center-cropped images, which is a common data augmentation method as in~{\cite{Han2015}}. Similarly, a comparative experiment with the method of Thinet is conducted on the ImageNet dataset.

\subsubsection{VGGNet and ResNet}
Our approach is evaluated on two popular convolution networks: VGGNet and ResNet. Specifically, VGG-16 model with batch normalization layer is used on all three datasets and ResNet-50 model with bottleneck structure is adopted on the CIFAR and SVHN datasets. Furthermore, FC layers of the VGG-16 model are replaced with a global average pooling (GAP) layer on the CIFAR datasets. 

\begin{figure*}[t]
	\centering
	\includegraphics[width=2.0\columnwidth]{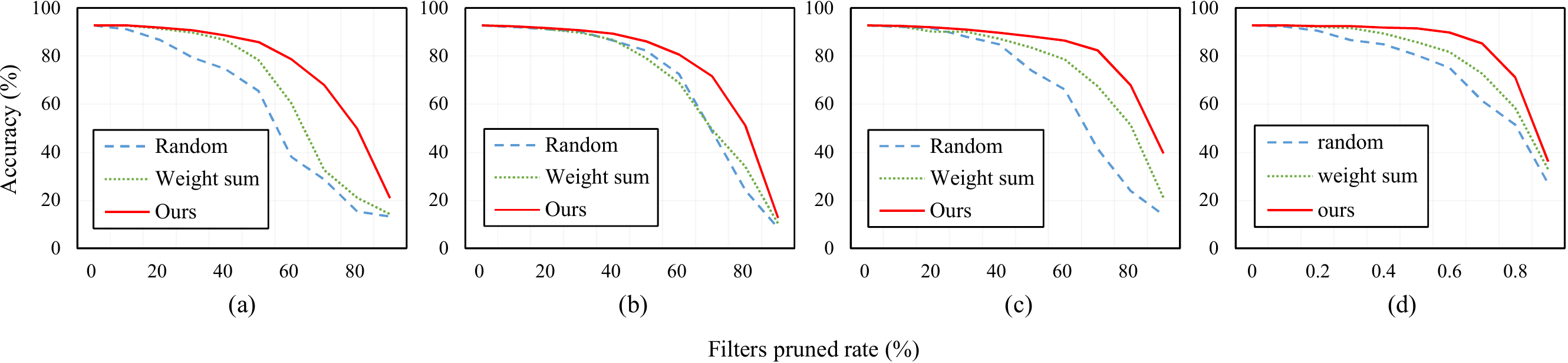}
	\vskip -2mm
	\caption{{single layer performance comparison with two simple channel selection strategies on the CIFAR-10 dataset. Three different layers of VGG-16 model are pruned respectively with different compression rate. Here, no fine-tuning is followed after pruning. (a) $\sim$ (d) represent conv2, conv4, conv5, conv7 respectively. }}
	\label{fig:singleLayer}
\end{figure*}

\begin{figure}[t]
	\centering
	\includegraphics[width=0.8\columnwidth]{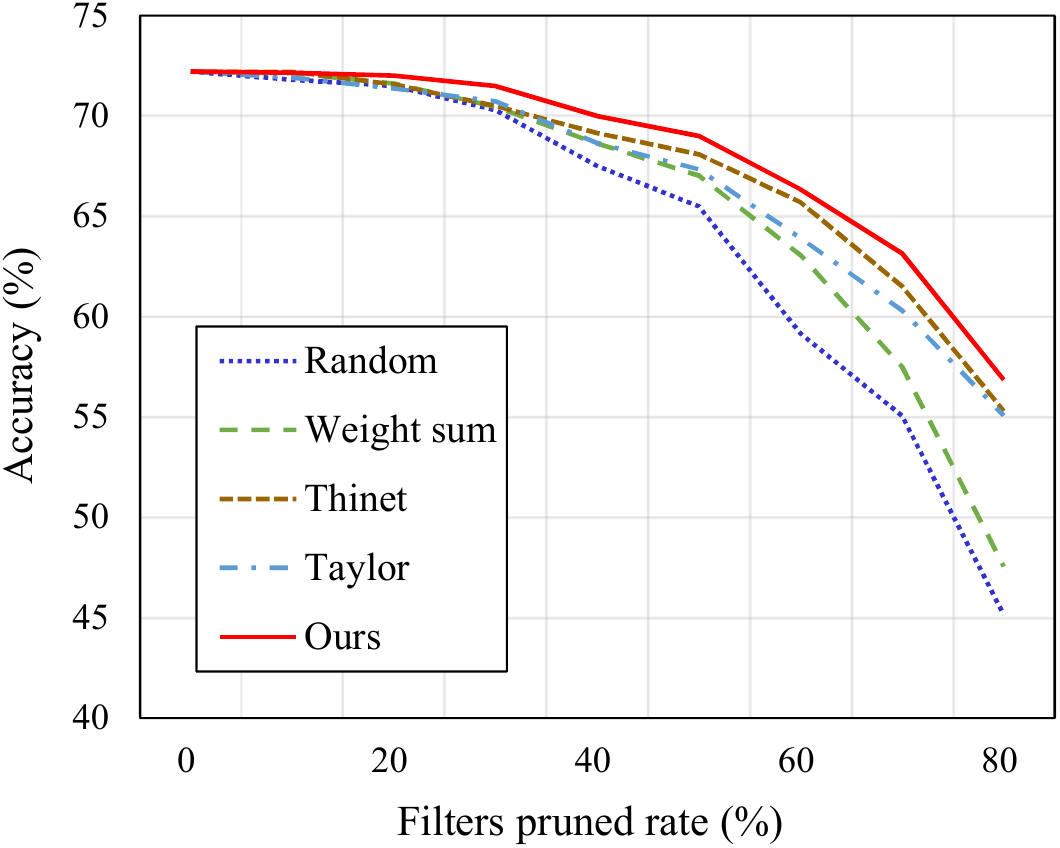}
	\vskip -2mm
	\caption{{performance comparison of several channel pruning methods on the CIFAR-100 dataset.}}
	\label{fig:wholeModel}
\end{figure}

\subsection{Implementation details}
\subsubsection{Training}
VGGNet and ResNet are trained from scratch as baselines. VGGNet is trained on all three datasets. ResNet is trained on the CIFAR and SVHN datasets. And These two networks are trained with batch size of 128 for 200 epochs on the CIFAR dataset and 50 epochs on the SVHN dataset. Meanwhile, The cross entropy loss is adopted as criterion function and Stochastic Gradient Descent (SGD) is took as optimization scheme. The initial learning rate for these two datasets is set to 0.1, and is multiplied by 0.1 at 40\% and 80\% of the total number of epochs respectively. In addition, weight decay is adopted to prevent networks from overfitting and momentum is used to accelerate convergence. On the ImageNet dataset, VGG-16 model is trained using batch size of 256 for 60 epochs. The initial learning rate remains unchanged, and decays by 10 at epoch 20 and 40. The same weight decay and momentum strategies as the previous datasets are adopted.

\subsubsection{Pruning}
For VGGNet, the network is pruned layer by layer sequentially. For ResNet, the channel numbers of each block need to be consistent due to the bottleneck structure, which make it difficult to prune the last layer of each block. Therefore, only first two layers of each block are pruned. Before pruning, the Hessian matrix of each layer need to be calculated . Here, the same data preprocessing operations as training stage are used. It is unnecessary to calculate the Hessian matrix using all images in training set, so only 1\% images of the entire training set are randomly chosen. Meanwhile, 10 volumes are randomly sampled at different spatial locations for each input image. Then these chosen samples constitute the subset of calculating the Hessian matrix. In this way, no more than 100,000 volumes are sampled even on the ImageNet dataset. We try different sampling strategies and find current sample number is enough. In Algorithm~{\ref{alg:genetic_algorithm}}, the population size is set to 20. And the maximum iteration number $T$ is set to 10 times of the channel number of the layer to be pruned. 

\subsubsection{Fine-tuning}
After pruning, a compact network with lower accuracy is obtained, which is then fine-tuned. On the CIFAR and SVHN datasets, the pruned model is fine-tuned for one epoch with $1.0\times10^{-3}$ learning rate after pruning each layer. On the ImageNet dataset, the pruned model is fine-tuned for two epochs to restore larger accuracy drop. After pruning the last layer, extra 20 epochs are adopted with learning rate from $1.0\times10^{-3}$ to $1.0\times10^{-5}$. In our experiments, all fine-tuning operations are based on knowledge distillation framework (as shown in Section \ref{KnowledgeDistill}). Without loss of generality, attention losses are placed at the last layer of each group for VGGNet and ResNet.

\subsection{Results and analysis}
\subsubsection{Single layer pruning}
\label{SingleLayerPruning}
In order to demonstrate the necessity of channel selection, our approach is compared with two simple channel selection strategies. The method of Random randomly selects a subset from all filters. The method of Weight sum removes the filters with weak activations which are determined by the absolute sum of filters. When our approach removes n channels at layer $i+1$, the same number of filters are removed at the $i$-th layer by Random and Weight sum methods. 

Fig. \ref{fig:singleLayer} summarizes the results on the CIFAR-10 dataset. As expected, our approach shows stronger ability of preserving accuracy and better robustness than other methods at different layers with different compression rate. The reason comes in two parts, one is that correlations of different channels is took into consideration in our algorithm, and the other is that our pruning criterion is closely linked with the final loss. It is worth noticing that the method of Weight sum performs even worse than the method of Random. In fact, there is no direct correlations between weights magnitude and the loss function, thus the method of Weight sum may remove some important filters. We also observed that the method of Random is not robust enough and it displays significant fluctuation. 

\subsubsection{Whole model pruning}
Our method is compared with four existing channel selection strategies: Thinet, Taylor \cite{Molchanov2016}, Random, and Weight sum. First 12 layers of VGG-16 model are pruned with different compression rate respectively. Fig. \ref{fig:wholeModel} summarizes the results on the CIFAR-100 dataset. Obviously, our approach achieves better performance than other methods. Due to the random nature of genetic algorithm, our approach are repeated 3 times and then the averaged results are reported. Next, the performance of each method will be summarized successively. 
\begin{table*}[t]
	\small
	\renewcommand\arraystretch{1.2}
	\caption{The performance of our approach in different data domains. VGG-16 and ResNet-50 models are pruned on the CIFAR and SVHN datasets with different compression rate. Here, M means million ($10^6$); $\downarrow$ represent the decrement of parameters or FLOPs compared with the baseline models.}
	\label{table:SeveralDatasets}
	\begin{center}
		\begin{tabular}{ccccccc}
			\toprule
			\makecell[c]{Models} & \makecell[c]{Datasets} & \makecell[c]{Test error} & \makecell[c]{Parameters} & \makecell[c]{Parameters $\downarrow$} &\makecell[c]{FLOPs} & \makecell[c]{FLOPs $\downarrow$}\\ 
			\midrule
			\makecell[c]{VGG-16 (Baseline)}& SVHN & 3.97\% & 14.7M & - & $6.26\times 10^8$ & -\\
			\makecell[c]{VGG-16 (58\% pruned)}& SVHN &  \textbf{3.87\%} & 1.71M & 88.4\% & $2.10\times 10^8$ & 66.4\%\\
			\makecell[c]{ResNet-50 (Baseline)}& SVHN &  3.56\% & 25.7M & - & $3.18\times 10^9$ & -\\
			\makecell[c]{ReseNet-50 (16\% pruned)}& SVHN &  \textbf{3.48\%} & 9.05M & 64.8\% & $1.74\times 10^9$ & 45.3\%\\
			\midrule
			\makecell[c]{VGG-16 (Baseline)}& CIFAR-100 & 27.79\% & 12.8M & - & $6.10\times 10^8$ & -\\
			\makecell[c]{VGG-16 (37\% pruned)}& CIFAR-100 &  \textbf{27.99\%} & 4.5M & 64.8\% & $3.82\times 10^8$ & 37.4\%\\
			\makecell[c]{ResNet-50 (Baseline)}& CIFAR-100 & 25.71\% & 25.7M & - & $3.18\times 10^9$ & -\\
			\makecell[c]{ReseNet-50 (16\% pruned)}& CIFAR-100 & \textbf{25.90\%} & 9.24M & 64.0\% & $1.74\times 10^9$ & 45.3\%\\
			\midrule
			\makecell[c]{VGG-16 (Baseline)}& CIFAR-10 & 7.29\% & 14.7M & - & $6.26\times 10^8$ & -\\
			\makecell[c]{VGG-16 (52\% Pruned)}& CIFAR-10 &\textbf{7.26\%} & 2.35M & 84.0\% & $2.74\times 10^8$ & 56.2\%\\
			\makecell[c]{ResNet-50 (Baseline)}& CIFAR-10 & 6.34\% & 25.7M & - & $3.18\times 10^9$ & -\\
			\makecell[c]{ReseNet-50 (20\% pruned)}& CIFAR-10 & \textbf{5.85\%} & 6.44M & 74.9\% & $1.62\times 10^9$ & 49.1\%\\
			\bottomrule
		\end{tabular}
	\end{center}
\end{table*}

\begin{table}[t]
	\small
	\caption{Performance of two channel pruning methods on the ImageNet dataset.}
	\label{table:ImageNet_thinet}
	\begin{center}
		\begin{tabular}{ccccc}
			\toprule
			\makecell[c]{Method} & \makecell[c]{Top-1} & \makecell[c]{Top-5} & \makecell[c]{Parameters $\downarrow$} & \makecell[c]{FLOPs $\downarrow$}\\ 
			\midrule
			\makecell[c]{Thinet}& +1.46\% & +1.09\% & 4.90\% & 69.03\%\\
			\makecell[c]{ours}& +2.08\% & +1.05\% & 5.52\% & 69.32\%\\
			\bottomrule
		\end{tabular}
	\end{center}
\end{table}

\begin{table}[t]
	\small
	\caption{Test error of two fine-tuning frameworks on the CIFAR-100 dataset. Here, increased error means the test error increment for the pruned model compared with the baseline model.}
	\label{table:KnowledgeDistill}
	\begin{center}
		\begin{tabular}{cccc}
			\toprule
			\makecell[c]{Models} & \makecell[c]{Compression rate} & \makecell[c]{Test error} & \makecell[c]{Increased error}\\
			\midrule
			\makecell[c]{FT}& 40\% & 30.01\% & 2.22\% \\
			\makecell[c]{FT-KD}& 40\% & 29.18\% & 1.39\% \\
			\midrule
			\makecell[c]{FT}& 80\% & 43.14\% & 15.35\%\\
			\makecell[c]{FT-KD}& 80\% & 39.97\% & 12.18\% \\
			\bottomrule
		\end{tabular}
	\end{center}
\end{table}

The method of Thinet minimizes Eq.~({\ref{eq:layer_wise_error_function}}) without approximation via greedy algorithm. As shown in Fig. \ref{fig:wholeModel} , our approach obtains a more competitive result than it especially for relatively high compression rate. Experimental results demonstrate that genetic algorithm is able to discover a better solution for relatively large solution space. Meanwhile, our method is relatively high efficiency, since it only needs to traverse the entire training set only once when performing single layer pruning.

The method of Taylor associates pruning criterion with the final loss, and prunes filters which have little effect on loss function. To accelerate the pruning process, it adopts the first-order Taylor expansion to approximate the loss change. It can be found that our approach outperforms the method of Taylor by a large margin. It focuses more on model acceleration, thus takes the first-order Taylor expansion to roughly approximate the objective function, while the second-order terms is adopted to do that in our algorithm.

Similar as Section \ref{SingleLayerPruning} , Weight sum and Random methods still show poor performance, which makes it difficult to restore back to the original accuracy even with fine-tuning. For instance, the test error of these two methods are 10\% lower than other methods when 80\% filters are removed. It is discussed that data-free pruning methods such as Weight sum and Random methods are relatively coarse pruning strategies, and their performance is not nearly as good as data-dependent methods, e.g., the methods of Thinet and Taylor.

The performance of our method is also explored with VGG-16 model on the ImageNet dataset. According to the sensitivity analysis (described in Section \ref{PruningWholeModel}) , filters pruned rate for four group is set to 60\%, 50\%, 40\%, 20\% respectively. In total, over one half filters are removed except conv13. Conv13 is tightly associated with the final feature representation, thus it is reserved. Table~{\ref{table:ImageNet_thinet}} shows the performance of resulting model. It can be found that our method outperforms the method of Thinet in accuracy with similar parameters and FLOPs reduction. Because some redundant channels are removed, the pruned model even achieve better performance than the original VGG-16 model. However, two methods show only around 5\% parameters reduction, since almost 89\% of parameters for VGG-16 model are distributed in FC layers. At present, replacing FC layers with a GAP layer is a common way of achieving high compression rate. 

To further explore the limits of our approach, VGGNet and ResNet are pruned with different compression rate on the CIFAR and SVNH datasets respectively, to verify the performance of the proposed method in different data domains. As shown in Table \ref{table:SeveralDatasets} , 50\% and 60\% channels of VGG-16 model are removed on the CIFAR-10 and SVHN datasets respectively, yet the performance of resulting models is even better than the baseline models. The pruned VGGNet achieves 8$\times$ parameters compression and 3$\times$ FLOPs reduction. Moreover, we found ResNet is more difficult to be compressed than VGGNet. This indicates that ResNet has lower redundancy than VGGNet, since the bottleneck structure of ResNet stops some layers from being pruned. We observed that the compression rate on the CIFAR-100 dataset is lower than other datasets. For instance, VGG-16 model shows higher test error on the CIFAR-100 dataset with 37\% compression rate. The reason for this is most probably that CIFAR-100 contains more classes.

\subsubsection{Fine-tune based on knowledge distillation framework}
The fine-tuning strategy based on knowledge distillation framework (FT-KD) is compared with the baseline (FT). As shown in Table \ref{table:KnowledgeDistill} , the VGG-16 model is pruned on the CIFAR-100 dataset with two different compression rate. We observed that the test error of FT-KD are 1\% and 4\% lower than FT with the compression rate of 40\% and 80\% respectively. This indicates that knowledge distillation framework by transferring knowledge from the original model into the resulting model is actually an effective way of fine-tuning.

\begin{figure}[t]
	\centering
	\includegraphics[width=0.75\columnwidth]{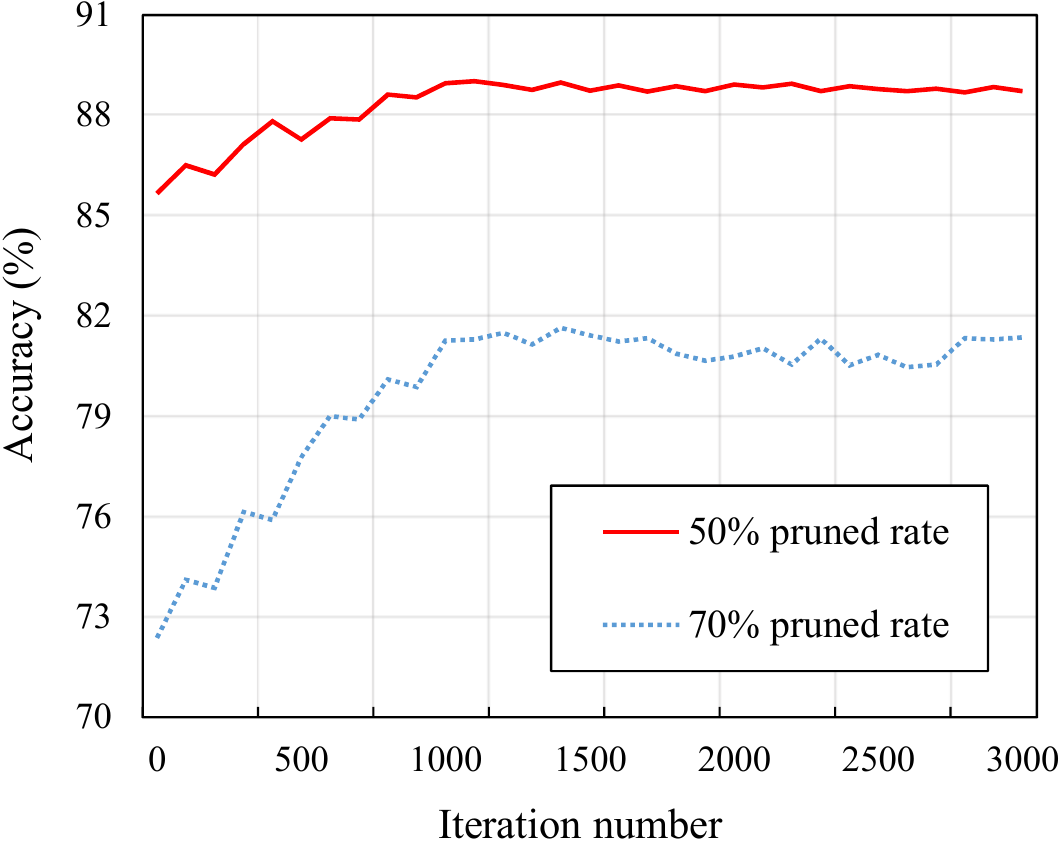}
	\vskip -2mm
	\caption{{the robustness evaluation of genetic algorithm. 6-th layer of VGG-16 model is pruned on the CIFAR-10 dataset with 50\% and 70\% compression rate respectively.}}
	\label{fig:iteration}
\end{figure}

\subsubsection{The robustness of genetic algorithm}
We first randomly selected one layer from 13 convolution layers of VGG-16 model, and then performed single layer pruning with two different compression rate. As shown in Fig.~{\ref{fig:iteration}} , with the increase of the iteration number, our algorithm begins to converge gradually after 1,000 generations. It can be found that little volatility still exists after 1200 generations due to the random nature of genetic algorithm. But the volatility has little influence on results, since the following fine-tuning process will compensate the accuracy drop after pruning.

\section{Conclusion and future works}
In this paper, a novel channel pruning method based on genetic algorithm is proposed, which achieves state-of-the-art results on the CIFAR-10 and ImageNet datasets. Firstly, a two-step approximation fitness function is designed, which makes the pruning process more efficient. secondly, a common way of fine-tuning based on knowledge distillation framework is presented. Thirdly, the proposed method has been verified on three benchmark datasets with two popular CNN models. Experimental results demonstrate the effectiveness of our approach.  

At present, most model compression and acceleration methods are designed for image classification, yet few researchers pay attention to other computer vision tasks such as object detection. One major reason is that deep neural networks for object detection depend more on complex feature presentation, so that these networks are very sensitive to model compression. In the future, we plan to design network compression scheme for object detection. Furthermore, we would like to combine the existing pruning strategies with other network compression strategies to explore more compact model with less accuracy drop.





\section*{Acknowledgment}
This work was partly supported by $\times\times$.

\end{document}